\pdfoutput=1
\documentclass[11pt]{article}
\usepackage[]{EACL2023}

\usepackage{times}
\usepackage{latexsym}
\usepackage[T1]{fontenc}
\usepackage[utf8]{inputenc}
\usepackage{microtype}
\usepackage{inconsolata}
\usepackage{url}
\usepackage{amsmath,amssymb,amsfonts}
\usepackage{textcomp}
\usepackage{xcolor}
\usepackage{mathtools}
\usepackage{subfigure}
\usepackage{multirow,booktabs}

\usepackage{amsmath,amsfonts,bm}

\def\eqref#1{equation~\ref{#1}}

\def\1{\bm{1}}

\def\vt{{\bm{t}}}

\def\vv{{\bm{v}}}

\def\vx{{\bm{x}}}

\def\mI{{\bm{I}}}

\def\mM{{\bm{M}}}

\def\mT{{\bm{T}}}

\def\mV{{\bm{V}}}

\DeclareMathAlphabet{\mathsfit}{\encodingdefault}{\sfdefault}{m}{sl}
\SetMathAlphabet{\mathsfit}{bold}{\encodingdefault}{\sfdefault}{bx}{n}

\def\gL{{\mathcal{L}}}

\newcommand{\R}{\mathbb{R}}

\DeclareMathOperator*{\transformerenc}{Trans-Enc}
\DeclareMathOperator*{\relu}{ReLU}
\DeclareMathOperator*{\linearproj}{Linear-Projection}
\DeclareMathOperator*{\upsampling}{Up-Sampling}
\DeclareMathOperator*{\downsampling}{Down-Sampling}

\title{Improving Cross-modal Alignment for Text-Guided Image Inpainting}

\author{Yucheng Zhou, Guodong Long \\
         Australian AI Institute, School of Computer Science, FEIT, University of Technology Sydney \\
         {\tt yucheng.zhou-1@student.uts.edu.au, guodong.long@uts.edu.au}\\
         }

\begin{document}
\maketitle
\begin{abstract}
Text-guided image inpainting (TGII) aims to restore missing regions based on a given text in a damaged image. Existing methods are based on a strong vision encoder and a cross-modal fusion model to integrate cross-modal features. However, these methods allocate most of the computation to visual encoding, while light computation on modeling modality interactions. Moreover, they take cross-modal fusion for depth features, which ignores a fine-grained alignment between text and image. Recently, vision-language pre-trained models (VLPM), encapsulating rich cross-modal alignment knowledge, have advanced in most multimodal tasks. In this work, we propose a novel model for TGII by improving cross-modal alignment (CMA). CMA model consists of a VLPM as a vision-language encoder, an image generator and global-local discriminators. To explore cross-modal alignment knowledge for image restoration, we introduce cross-modal alignment distillation and in-sample distribution distillation. In addition, we employ adversarial training to enhance the model to fill the missing region in complicated structures effectively. Experiments are conducted on two popular vision-language datasets. Results show that our model achieves state-of-the-art performance compared with other strong competitors. 
\end{abstract}

\section{Introduction}
\begin{figure}[t]
    \centering
    \includegraphics[width=0.85\linewidth]{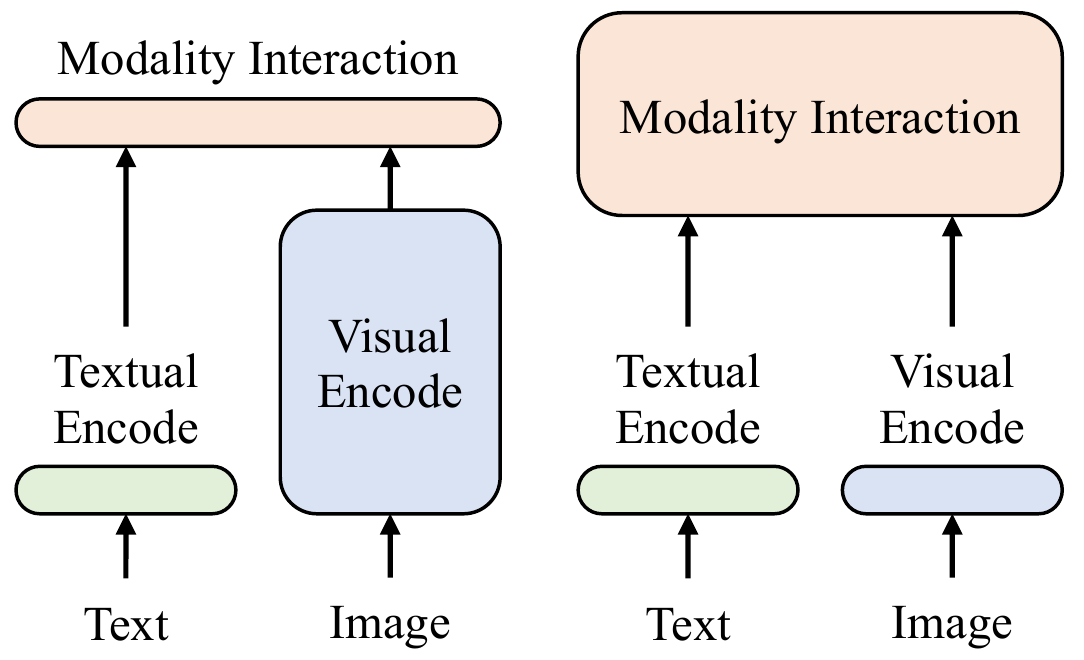}
    \caption{Different categories of vision-and-language models. \textit{left}: most of the computation on visual encoding; \textit{right}: most of the computation on modeling modality interactions.}
    \label{fig:intro}
\end{figure}
Text-guided image inpainting (TGII), involving computer vision (CV) and natural language processing (NLP), aims to restore visual content for a missing area in a damaged image based on a given text \cite{Zhang20Text1}. With the development of CV and NLP, it plays an essential role in many real-world applications, such as image editing \cite{Zhu20In}, damaged image restoration \cite{Liu19Coherent}, and image rendering \cite{Kirillov20PointRend}. Therefore, it has become one of the most crucial areas in CV and NLP.

Existing methods \cite{Zhang20Text,Lin20MMFL,Wu21Adversarial} adopt an encoder-decoder framework as their backbone with a vision-language fusion module to introduce textual information. These methods use separate encoders for images and texts, heavier on the former. Then, the vision-language fusion module is used to integrate the features from the two modalities through a simple similarity calculation of features, or shallow attention layers \cite{vaswani2017attention}, as shown on the left of Figure~\ref{fig:intro}. Most of the computation of these methods on visual encoding, while textual features only serve as a complement to deep visual features, which ignores the importance of deep interaction of multimodal information. Moreover, these methods share a common drawback: they do not perform well for natural image datasets with a wide variety of objects (e.g., MSCOCO \cite{lin2014microsoft}). The reason is that these methods lack fine-grained alignment knowledge of texts and images to guide the fusion of cross-modal information in multimodal interactions. Besides, their fusion modules lack powerful cross-modal reasoning capabilities.

Recently, providing the success of pre-trained language/vision Transformer \cite{Devlin19BERT,Dosovitskiy21Image,Zhou22EventBERT}, some works (e.g., ViLT \cite{Kim21ViLT} and SimVLM \cite{Wang21SimVLM}) pre-train a vision-and-language Transformer on a large-scale image-text dataset. These Vision-and-Language Pre-trained Models (VLPM) achieve exciting performance on many multimodal downstream tasks. The reason is that VLPM encapsulates rich image and text alignment knowledge and has a strong cross-modal reasoning capability \cite{Chen20UNITER,Kim21ViLT}. To enhance the cross-modal interaction between images and texts, \citet{Kim21ViLT} propose ViLT, which focuses most of the computation on modeling the multimodal interaction, as shown on the right side of Figure~\ref{fig:intro}.

Motivated by \citet{Kim21ViLT}, we propose a novel model enhanced by cross-modal alignment (CMA) for text-guided image inpainting, comprising a vision-and-language encoder, an image generator and global-local discriminators. Different from previous works \cite{Zhang20Text,Lin20MMFL,Wu21Adversarial}, we employ a vision-and-language encoder based on VLPM to encode images and texts instead of separate vision and language encoders. The vision-and-language encoder can encode images and texts in a cross-modal interaction manner to implement visual priors reconstruction. Then, the visual features integrating textual information (i.e., reconstructed visual priors) obtained by VLPM are passed to an image generator to generate a restored image. To improve cross-modal alignment for image inpainting, we introduce cross-modal alignment distillation to guide a fine-grained fusion of cross-modal knowledge. Moreover, to further strengthen the visual priors reconstruction, we utilize in-sample distillation to enhance the model's cross-modal reasoning capability for the content of missing regions. Besides, we employ adversarial learning to improve the quality of generated images through global-local discriminators.

Experiments are conducted on two popular image-text datasets with a wide variety of objects (i.e., MSCOCO \cite{lin2014microsoft} and Flicker30K \cite{Plummer17Flickr30k}). Experimental results demonstrate that our method achieves state-of-the-art performance compared to other strong competitors. In addition, we analyze the effectiveness of each module of our method and the impact of cross-modal alignment. Moreover, we also conduct extensive analysis to verify the effectiveness of our method.

\section{Method}
\begin{figure*}[t]
    \begin{center}
    \includegraphics[width=0.9\linewidth]{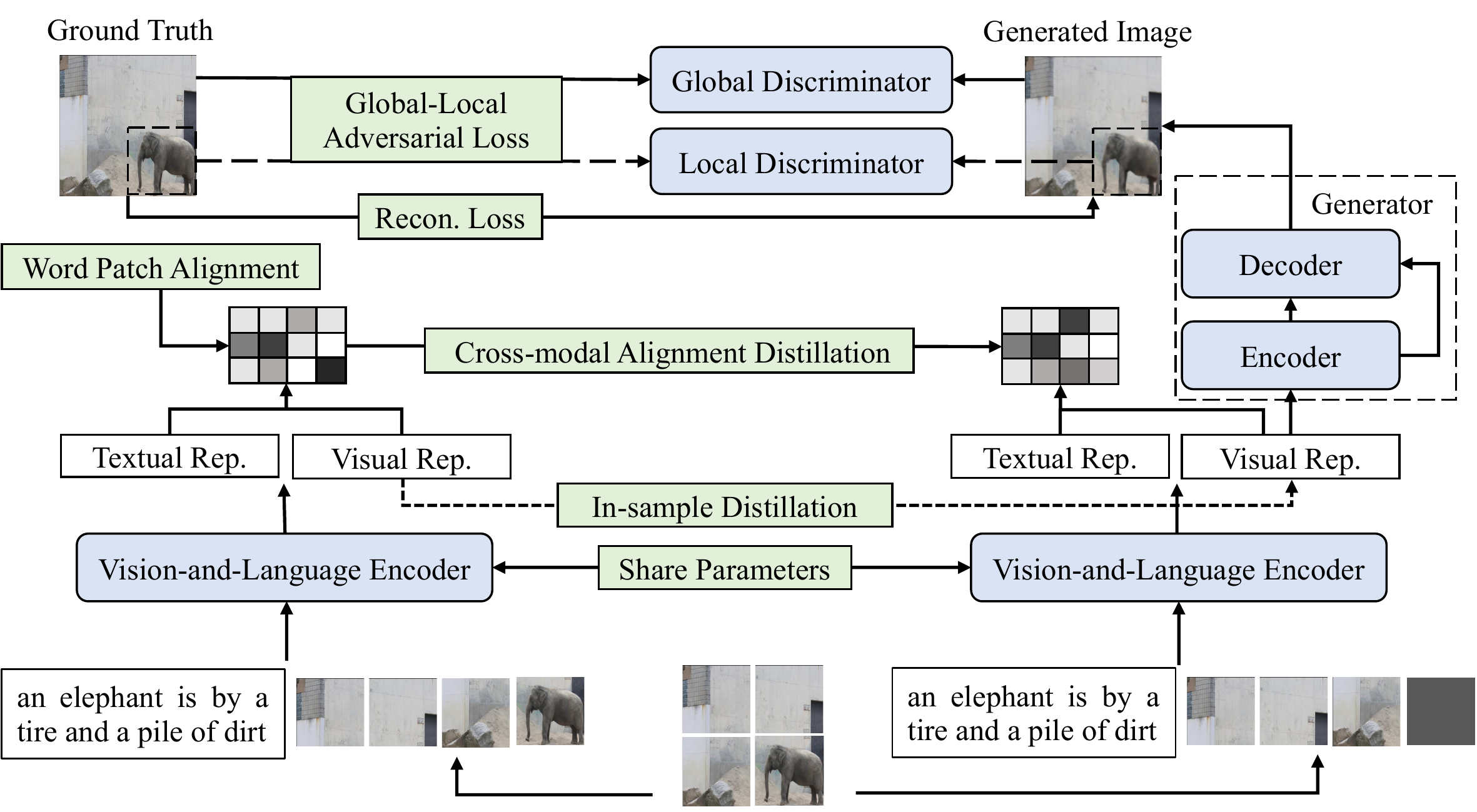}
    \end{center}
    \caption{Overview of our model. Blue rounded rectangles denote trainable modules. Green right rectangles indicate training objectives or operations.}
    \label{fig:model}
\end{figure*}
In this section, we will introduce our CMA model as shown in Figure~\ref{fig:model}. CMA model consists of a vision-and-language encoder, an image generator and global-local discriminators. In addition, details about training and inference are elaborated.

\subsection{Vision-and-Language Encoding}
In previous works \cite{Zhang20Text,Lin20MMFL,Wu21Adversarial}, images and texts are encoded by separate encoders. Specifically, a CNN-based image encoder is used to extract visual features for images, while a RNN-based text encoder is used to encode text to obtain textual features. Next, the image and text features are integrated by a multimodal fusion module to obtain multimodal representations (i.e., reconstructed visual priors). In this work, we employ a novel Transformer-based cross-modal encoder for image and text encoding, a vision-and-language encoder instead of two separate encoders. For language encoding, we employ a word embedding matrix to embed text $\mT$ $\in$ $\R^{L \times |V|}$ into $\bar \mT$ $\in$ $\R^{L \times e}$, where $L$, $|V|$, and $e$ denote the length of text, size of vocabulary and size of embedding, respectively. For visual encoding, an image $\mI$ $\in$ $\R^{C \times H \times W}$ is sliced into patches and flattened to $\mV$ $\in$ $\R^{N \times (P^2 \times C)}$,  where $P$ is the size of patch and $N$ equal to $(H \times W)/ P^2$. Following \citet{Kim21ViLT}, $\mV$ is embedded into $\bar \mV$ $\in$ $\R^{N \times e}$: 
\begin{align}
    \bar \mV = \linearproj(\mV)
    \label{equ:linear}
\end{align}
where the details of linear projection can be found in \cite{Kim21ViLT}. Next, images and texts are encoded in a cross-modal interaction manner:
\begin{align}
    [\hat \mT;\hat \mV] = \transformerenc([\bar \mT;\bar \mV])
    \label{equ:enc}
\end{align}
where $[;]$ denote a concatenation operation, and outputs of the vision-and-language encoder can be represented as $\hat \mT$ $\in$ $\R^{L \times e}$ for textual representations and $\hat \mV$ $\in$ $\R^{N \times e}$ for reconstructed visual priors. Compared to CNN constrained by its inherent properties (e.g., spatial-invariant kernels), which are not conducive to understanding global features \cite{Wan21High}, Transformer-based encoders have a natural advantage in encoding global features across modalities.

\subsection{Image Generation}
The process of image generation includes two stages. The first stage is to downsample the visual priors to extract deep visual representations. Next, performing upsampling for deep visual representations to generate a restored image. Different from the previous works \cite{Zhang20Text,Lin20MMFL,Wu21Adversarial}, we do not integrate a fusion module to introduce textual features in the image generation process because the texts are introduced into the vision-and-language encoders. Although transformers demonstrate their effectiveness in long-term relations and the advantage of understanding global features, their computational complexity is quadratic with the input length, which hinders image generation \cite{Wan21High}. Therefore, our image generator consists of two CNN-based components, an encoder for downsampling and a decoder for upsampling. Similar to \cite{Zhang20Text1}, we employ a 5-layer ResNet \cite{He16Deep} as the downsampling encoder, and the visual priors are fed into it to obtain deep visual representations:
\begin{align}
    \vv = \downsampling(\hat \mV) \label{equ:ds}
\end{align}
In addition, we pass $\vv$ into the upsampling decoder to perform image reconstruction to generate a restored image, and the upsampling decoder consists of a 5-layer residual generator network. 
\begin{align}
    \mI_r = \upsampling(\vv, \hat \mV) \label{equ:us}
\end{align}
The visual priors obtained by the vision-and-language encoder are input to the upsampling decoder through a skip connection to provide detailed information that forgets in the downsampling stage.

\subsection{Training}
For model training, we construct five training objectives, including cross-modal alignment distillation, in-sample distillation, word patch alignment, reconstruction loss and global-local adversarial loss.

\paragraph{Cross-Modal Alignment Distillation. }
To guide image inpainting through cross-modal alignment knowledge, we pass the original image and text to the vision-and-language encoder to obtain a text representation $\hat \mT_o$ and a visual priors $\hat \mV_o$. Then, we obtain the correlation map $\mM_o$ between each token of text and each image patch by calculating the similarity between them. The correlation map corresponding to the corrupted image and text is denoted as $\mM_r$. Next, we compute the pair-wise similarity distillation loss between the two correlation maps:
\begin{align}
\ell_{\mathrm{CMAD}} &= \frac{1}{N \times L} \sum_{i = 1}^{L} \sum_{j = 1}^{N} (a_{i j}^{r}-a_{i j}^{o})^{2} \\
a_{i j}^r &= \frac{\vt_i^{r\top} \vv_j^r}{\|\vt_i^r \|_{2} \|\vv_j^r \|_{2}}, a_{i j}^r \in \mM_r\\
a_{i j}^o &= \frac{\vt_i^{o\top} \vv_j^o}{ \|\vt_i^o  \|_{2}  \|\vv_j^o  \|_{2}}, a_{i j}^o \in \mM_o
\end{align}
where $\vt_i^r$ $\in$ $\hat \mT$, $\vv_j^r$ $\in$ $\hat \mV$, $\vt_i^o$ $\in$ $\hat \mT_o$ and $\vv_j^o$ $\in$ $\hat \mV_o$, respectively.

\paragraph{In-Sample Distillation. }
Besides using cross-modal alignment knowledge to guide image inpainting, we propose in-sample distillation. The purpose of this objective is to guide the damaged image to infer visual priors closer to that from the original image based on the text and known regions, i.e.,
\begin{align}
    \ell_{\mathrm{ISD}} =\frac{1}{N} \sum_{i = 1}^N \mathrm{KL}\left(\vv_i^o \| \vv_i^r\right).
\end{align}
where $\mathrm{KL}$ denotes Kullback–Leibler divergence.

\paragraph{Word Patch Alignment. }
To maintain that the cross-modal alignment knowledge of the model is not degraded, we use the word patch alignment objective to preserve the cross-modal alignment knowledge integrated into the model.
\begin{align}
    \ell_{\mathrm{WPA}} &= \min _{\mathbf{T} \in \Pi(\mathbf{a}, \mathbf{b})} \sum_{i=1}^{N} \sum_{j=1}^{L} \mathbf{T}_{i j} \cdot c\left(\vv_i^o, \vt_j^o\right) \\
    c(\vv_i^o, \vt_j^o) &= 1-\frac{\vv_i^{o\top} \vt_j^o}{\|\vv_i^o \|_{2} \|\vt_j^o \|_{2}}\\
    \Pi(\mathbf{a}, \mathbf{b}) &= \{\mathbf{T} \in \mathbb{R}^{N \times L} \mid \mathbf{T} \mathbf{1}_{m}=\mathbf{a}, \notag \\
    & ~~~~~~~~~~~~~~~~~~~~~~~~~~~~~\mathbf{T}^{\top} \mathbf{1}_{n}=\mathbf{b}\}
\end{align}
where $\mathbf{1}_{n}$ denotes an n-dimensional all-one vector; $\mathbf{T}$ is a transport plan. The details can be found in \cite{Kim21ViLT}.

\paragraph{Reconstruction Loss. }
We adopt the $\ell_{1}$-norm error as the loss function between the restored image and its corresponding ground-truth image $\mI_{gt}$:
\begin{align}
 \ell_{\mathrm{1}} &= || \mI_{r} - \mI_{gt} ||_1,
\end{align}

\paragraph{Global Adversarial Loss. }
In image generation, the adversarial loss \cite{goodfellow2014generative} effectively improves the quality of the generated image. However, adversarial training is unstable since keeping the balance between the generator and discriminator is difficult. To tackle this problem, \citet{arjovsky2017wasserstein} propose the Wasserstein generative adversarial nets (WGAN) that can improve the stability of adversarial learning. We employ a WGAN hinge loss as the adversarial loss, and it can be formulated as follows:
\begin{align}
    \ell_{\mathrm{G-adv,G}} = & \mathbb{E}_{\hat{\vx} \sim P_{data}(\mI_r)}[-D(\hat{\vx})] \\
\notag \ell_{\mathrm{G-adv,D}} = & \mathbb{E}_{\vx \sim P_{data}(\mI_{gt})}[\relu(1-D(\vx))]\\
              +& \mathbb{E}_{\hat{\vx} \sim P_{data}(\mI_r)}[\relu(1+D(\hat{\vx}))]
\end{align}
where $D(\cdot)$ denotes a global discriminator; $\relu$ is the rectified linear unit function; $\gL_{adv,G}$ and $\gL_{adv,D}$ are loss function for inpainting model and discriminator, respectively.

The discriminator $D(\cdot)$ consists of 5-layer ResNet, followed by a fully connected layer. Each convolutional layer in ResNet applies spectral normalization to satisfy the Lipschitz constraints of Wasserstein GANs. 

\paragraph{Local Adversarial Loss. }
For the local discriminator, we use the same settings as the global discriminator, and the loss is defined as:
\begin{align}
    \ell_{\mathrm{L-adv,G}} = & \mathbb{E}_{\hat{\vx_l} \sim P_{data}(\mI_r)}[-D_l(\hat{\vx_l})] \\
\notag \ell_{\mathrm{L-adv,D}} = & \mathbb{E}_{\vx_l \sim P_{data}(\mI_{gt})}[\relu(1-D_l(\vx_l))]\\
              +& \mathbb{E}_{\hat{\vx_l} \sim P_{data}(\mI_r)}[\relu(1+D_l(\hat{\vx_l}))]
\end{align}
where $x_l$ denotes the restored image or ground truth corresponding to the missing region; $D_l(\cdot)$ denotes a local discriminator.

Considering the loss functions above, the objective function of our model in the generation stage can be defined as:
\begin{align}
    \ell_G =& \lambda \ell_{\mathrm{CMAD}} + \lambda \ell_{\mathrm{ISD}} + \alpha \ell_{\mathrm{WPA}} + \beta \ell_{\mathrm{1}} \notag \\
            +& \gamma \ell_{\mathrm{G-adv,G}} + \gamma \ell_{\mathrm{L-adv,G}} \label{equ:loss}
\end{align}
where the $\lambda$, $\alpha$, $\beta$ and $\gamma$ are the hyper-parameters used to balance the objective function.

In addition, the objective function of our model in the discriminative stage can be defined as:
\begin{align}
    \ell_D = \gamma \ell_{\mathrm{G-adv,D}} + \gamma \ell_{\mathrm{L-adv,D}} \label{equ:loss_d}
\end{align}

\subsection{Inference}
During inference, we first pass image $\mI$ and text $\mT$ into the cross-modal encoder (i.e., Equ.\ref{equ:linear} and Equ.\ref{equ:enc}) to obtain the visual representation $\hat \mV$. Next, we deliver the visual representation $\hat \mV$ into the generator (i.e., Equ.\ref{equ:ds} and Equ.\ref{equ:us}) to generate a restored image $\mI_{r}$.

\section{Experiments}
\begin{table*}[t]\small
    \centering
    \begin{tabular}{lcccccc}\toprule
    \multicolumn{1}{c}{\textbf{Method}} & \textbf{$\ell_{1}$ (\%)} $\downarrow$ & \textbf{FID} $\downarrow$  & \textbf{KID} $\downarrow$  & \textbf{TV loss (\%)} $\downarrow$ & \textbf{PSNR} $\uparrow$  & \textbf{SSIM (\%) $\uparrow$} \\\midrule
    CSA \cite{Liu19Coherent}                               & 5.14            & 50.88          & 2.85          & 4.32                 & 19.87          & 82.53             \\
    PICNet \cite{Zheng19Pluralistic}                             & 5.63            & 53.57          & 3.19          & 4.55                 & 19.58          & 81.87             \\
    CTSDG \cite{Guo21Image}                              & 5.03            & 48.95          & 2.56          & 4.31                 & 19.98          & 82.69             \\
    MMFL \cite{Lin20MMFL}                               & 4.36            & 44.33          & 2.22          & 4.28                 & 20.73          & 82.92             \\
    TGII \cite{Zhang20Text}                               & 4.22            & 44.02          & 1.87          & 4.39                 & 20.87          & 83.07             \\
    TDANet \cite{Zhang20Text1}                             & 4.13            & 42.38          & 1.67          & 4.59                 & 20.91          & 83.34             \\
    ALMR \cite{Wu21Adversarial}                               & 4.17            & 43.43          & 1.79          & 4.27                 & 20.81          & 83.15             \\
    CMA (ours)                          & \textbf{3.78}   & \textbf{39.52} & \textbf{1.34} & \textbf{4.17}        & \textbf{22.07} & \textbf{85.18}    \\\bottomrule
    \end{tabular}
    \caption{Results on MSCOCO with the center mask.}
    \label{tab:coco_center}
\end{table*}
\begin{table*}[t]\small
    \centering
    \begin{tabular}{lcccccc}\toprule
    \multicolumn{1}{c}{\textbf{Method}} & \textbf{$\ell_{1}$ (\%)} $\downarrow$ & \textbf{FID} $\downarrow$  & \textbf{KID} $\downarrow$  & \textbf{TV loss (\%)} $\downarrow$ & \textbf{PSNR} $\uparrow$  & \textbf{SSIM (\%) $\uparrow$} \\\midrule
    CSA \cite{Liu19Coherent}                                & 8.79            & 53.49          & 3.65          & 4.85                 & 18.99          & 75.50             \\
    PICNet \cite{Zheng19Pluralistic}                             & 9.15            & 56.80          & 3.72          & 5.09                 & 18.78          & 74.98             \\
    CTSDG \cite{Guo21Image}                              & 8.69            & 51.64          & 3.36          & 4.82                 & 19.13          & 75.84             \\
    MMFL \cite{Lin20MMFL}                                & 7.59            & 47.15          & 2.91          & 4.53                 & 20.07          & 76.16             \\
    TGII \cite{Zhang20Text}                               & 7.53            & 46.73          & 2.82          & 4.59                 & 20.27          & 76.46             \\
    TDANet \cite{Zhang20Text1}                             & 7.48            & 45.30          & 2.45          & 4.70                 & 20.55          & 76.93             \\
    ALMR \cite{Wu21Adversarial}                               & 7.54            & 46.01          & 2.61          & 4.46                 & 20.35          & 76.50             \\
    CMA (ours)                          & \textbf{7.00}   & \textbf{42.23} & \textbf{2.01} & \textbf{4.30}        & \textbf{21.75} & \textbf{78.67}    \\\bottomrule
    \end{tabular}
    \caption{Results on MSCOCO with the object mask.}
    \label{tab:coco_object}
\end{table*}
\begin{table*}[!t]\small
    \centering
    \begin{tabular}{lcccccc}\toprule
    \multicolumn{1}{c}{\textbf{Method}} & \textbf{$\ell_{1}$ (\%)} $\downarrow$ & \textbf{FID} $\downarrow$  & \textbf{KID} $\downarrow$  & \textbf{TV loss (\%)} $\downarrow$ & \textbf{PSNR} $\uparrow$  & \textbf{SSIM (\%) $\uparrow$} \\\midrule
    CSA \cite{Liu19Coherent}                                & 4.99            & 50.76          & 2.78          & 4.14                 & 19.93          & 83.97             \\
    PICNet \cite{Zheng19Pluralistic}                             & 5.29            & 52.25          & 3.14          & 4.39                 & 19.70          & 82.21             \\
    CTSDG \cite{Guo21Image}                               & 4.78            & 48.66          & 2.54          & 4.10                 & 20.39          & 84.21             \\
    MMFL \cite{Lin20MMFL}                                & 4.13            & 43.92          & 2.12          & 4.26                 & 20.78          & 83.96             \\
    TGII \cite{Zhang20Text}                                & 4.11            & 43.34          & 1.73          & 4.29                 & 21.48          & 84.49             \\
    TDANet \cite{Zhang20Text1}                               & 3.92            & 41.46          & 1.54          & 4.16                 & 21.36          & 84.17             \\
    ALMR \cite{Wu21Adversarial}                               & 4.05            & 42.43          & 1.60          & 4.12                 & 21.22          & 83.42             \\
    CMA (ours)                          & \textbf{3.61}   & \textbf{38.30} & \textbf{1.29} & \textbf{4.00}        & \textbf{22.55} & \textbf{86.33}    \\\bottomrule
    \end{tabular}
    \caption{Results on Flickr30K with the center mask.}
    \label{tab:filcker_center}
\end{table*}
\begin{table*}[t]\small
    \centering
    \begin{tabular}{lcccccc}\toprule
    \multicolumn{1}{c}{\textbf{Method}} & \textbf{$\ell_{1}$ (\%)} $\downarrow$ & \textbf{FID} $\downarrow$  & \textbf{KID} $\downarrow$  & \textbf{TV loss (\%)} $\downarrow$ & \textbf{PSNR} $\uparrow$  & \textbf{SSIM (\%) $\uparrow$} \\\midrule
        CSA \cite{Liu19Coherent}                                 & 8.60            & 53.18          & 3.57          & 4.75                 & 19.21          & 75.82             \\
        PICNet \cite{Zheng19Pluralistic}                              & 9.01            & 56.59          & 3.64          & 4.99                 & 19.16          & 75.95             \\
        CTSDG \cite{Guo21Image}                               & 8.45            & 51.35          & 3.30          & 4.69                 & 19.22          & 76.73             \\
        MMFL \cite{Lin20MMFL}                                & 7.58            & 46.53          & 2.86          & 4.53                 & 20.34          & 76.42             \\
        TGII \cite{Zhang20Text}                                & 7.39            & 46.52          & 2.73          & 4.56                 & 20.82          & 76.82             \\
        TDANet \cite{Zhang20Text1}                              & 7.37            & 44.72          & 2.42          & 4.55                 & 21.04          & 77.43             \\
        ALMR \cite{Wu21Adversarial}                                 & 7.40            & 45.74          & 2.52          & 4.36                 & 20.56          & 76.57             \\
        CMA (ours)                          & \textbf{6.86}   & \textbf{41.28} & \textbf{1.91} & \textbf{4.25}        & \textbf{22.07} & \textbf{79.93}   \\\bottomrule
    \end{tabular}
    \caption{Results on Flickr30K with the object mask.}
    \label{tab:filcker_object}
\end{table*}
\begin{table*}[t]\small
    \centering
    \begin{tabular}{lcccccc}\toprule
    \multicolumn{1}{c}{\textbf{Method}} & \textbf{$\ell_{1}$ (\%)} $\downarrow$ & \textbf{FID} $\downarrow$  & \textbf{KID} $\downarrow$  & \textbf{TV loss (\%)} $\downarrow$ & \textbf{PSNR} $\uparrow$  & \textbf{SSIM (\%) $\uparrow$} \\\midrule
CSA \cite{Liu19Coherent}                                & 4.04            & 56.88          & 2.74          & 3.70                 & 20.03          & 81.12             \\
PICNet \cite{Zheng19Pluralistic}                             & 3.78            & 47.33          & 2.46          & 3.74                 & 20.16          & 82.04             \\
CTSDG \cite{Guo21Image}                              & 3.63            & 38.05          & 1.98          & 3.71                 & 20.73          & 82.64             \\
MMFL \cite{Lin20MMFL}                               & 3.42            & 29.79          & 1.39          & 3.59                 & 20.68          & 82.36             \\
TGII \cite{Zhang20Text}                               & 3.54            & 32.57          & 1.51          & 3.63                 & 20.55          & 81.89             \\
TDANet \cite{Zhang20Text1}                             & 3.57            & 30.82          & 1.49          & 3.61                 & 20.79          & 82.68             \\
ALMR \cite{Wu21Adversarial}                               & 3.32            & 15.78          & 0.52          & 3.52                 & 20.66          & 80.61             \\
CMA (ours)                          & \textbf{3.07}   & \textbf{14.21} & \textbf{0.41} & \textbf{3.51}        & \textbf{20.84} & \textbf{82.68}    \\\bottomrule
\end{tabular}
\caption{Results on CUB with the center mask.}
\label{tab:cub_center}
\end{table*}
\begin{table*}[!t]\small
    \centering
    \begin{tabular}{lcccccc}\toprule
    \multicolumn{1}{c}{\textbf{Method}} & \textbf{$\ell_{1}$ (\%)} $\downarrow$ & \textbf{FID} $\downarrow$  & \textbf{KID} $\downarrow$  & \textbf{TV loss (\%)} $\downarrow$ & \textbf{PSNR} $\uparrow$  & \textbf{SSIM (\%) $\uparrow$} \\\midrule
CSA \cite{Liu19Coherent}                                & 6.76            & 59.91          & 3.00          & 4.21                 & 19.03          & 70.97             \\
PICNet \cite{Zheng19Pluralistic}                             & 6.42            & 50.16          & 2.82          & 4.32                 & 18.96          & 70.27             \\
CTSDG  \cite{Guo21Image}                              & 5.83            & 40.19          & 2.23          & 4.13                 & 19.38          & 72.16             \\
MMFL \cite{Lin20MMFL}                                 & 4.63            & 32.00          & 2.13          & 3.79                 & 20.32          & 78.24             \\
TGII \cite{Zhang20Text}                                & 4.72            & 34.39          & 2.07          & 3.93                 & 20.10          & 78.65             \\
TDANet \cite{Zhang20Text1}                               & 4.71            & 33.10          & 2.03          & 3.72                 & 20.40          & 77.77             \\
ALMR \cite{Wu21Adversarial}                               & 4.51            & 26.33          & 1.66          & 3.71                 & 20.25          & 75.69             \\
CMA (ours)                          & \textbf{4.24}   & \textbf{18.96} & \textbf{1.49} & \textbf{3.64}        & \textbf{20.48} & \textbf{78.75}    \\\bottomrule
\end{tabular}
\caption{Results on CUB with the object mask.}
\label{tab:cub_object}
\end{table*}
\subsection{Dataset and Evaluation Metrics}
We conduct the experiment on two datasets: MSCOCO \cite{lin2014microsoft} and Flicker30K \cite{Plummer17Flickr30k}. For the MSCOCO and Flicker30K datasets, we split them following their original training, validation and test set. Following \citet{Zhang20Text1},  we also set the mask for an image in two types: center mask and object mask. A center mask refers to a square mask taking 50\% area in the center of an image. An object mask indicates masking an image based on the object boxes provided by every image. To evaluate the performance of our model and other methods on these datasets, we utilize the $\ell_{1}$ loss, Frechet Inception Distance (FID) \cite{Heusel17GANs}, Kernel Inception Distance (KID) \cite{Heusel17GANs}, total variation (TV) \cite{rudin1992nonlinear}, peak signal-to-noise ratio (PSNR) \cite{Fardo16PSNR} and structural similarity index (SSIM) \cite{Wang04Image} as metrics to report the results. $\ell_{1}$ measure the $\ell_{1}$ distance between restored and original images. FID and KID measure the quality of restored images based on human perception. PSNR and SSIM measure structural similarity between restored and original images and the $\ell_{2}$ distance.

\subsection{Experimental Settings}
For training images in MSCOCO and Flicker30K, we resize them to make their minimal height/width 256 and crop them based on size 256 $\times$ 256 at the center. During training, we set the $\lambda$ as 2, $\alpha$ as 1, $\beta$ as 1 and $\gamma$ as 0.1 in the objective function, and the model is trained by an Adam optimizer \cite{kingma2014adam} with the learning rate of $1 \times 10^{-4}$. In the vision-and-language encoder, the patch size, intermediate size and hidden size are 32, 3072 and 768. For a masked patch, we utilize a special token \texttt{[Vmask]} as input. We employ ViLT \cite{Kim21ViLT} to initialize our vision-and-language encoder. The weight decay and gradient clipping are set to 0.01 and 1.0. The maximum training epoch and batch size are 200 and 128. Warmup steps and maximum sequence length are set to 2000 and 40. Our experiments are conducted on 2 $\times$ V100 GPUs. We choose models with the best result on the validation set and report the results on the test set based on the models. 

\subsection{Main Results}
The experimental results of our method and previous works on MSCOCO and Flicker30K are shown in Table \ref{tab:coco_center}, Table \ref{tab:coco_object},  Table \ref{tab:filcker_center} and Table \ref{tab:filcker_object}. From the tables, we can see that our model achieves state-of-the-art results compared with other strong competitors. In addition, we can make two observations: Firstly, using text to guide image inpainting can significantly improve the performance. For example, the text-guided image inpainting methods (i.e., MMFL, TGII, TDANet, ALMR and CMA) outperform standard image inpainting methods (i.e., CSA, PICNet and CTSDG). Next, for the gap between results on center masks and object masks, it shows that recovering completely removed objects is more difficult. Besides, we show results on the CUB \cite{wah2011caltech} dataset without a wide variety of objects in Tabel~\ref{tab:cub_center} and Tabel~\ref{tab:cub_object}. Results show our method outperforms other methods.

\subsection{Ablation Study}
\begin{table}[t] \small
    \setlength\tabcolsep{5.2pt}
    \centering
    \begin{tabular}{lcccc}\toprule
    \multicolumn{1}{c}{\textbf{Method}} & \textbf{FID} $\downarrow$   & \textbf{KID} $\downarrow$  & \textbf{PSNR} $\uparrow$  & \textbf{SSIM (\%)} $\uparrow$ \\\midrule
    CMA                                 & 39.52          & 1.34          & 22.07          & 85.18             \\\midrule
    w/o G-Adv.                          & 43.72          & 2.17          & 21.39          & 84.12             \\
    w/o L-Adv.                          & 42.59          & 1.84          & 21.74          & 84.75             \\
    w/o Adv.                            & 50.94          & 2.91          & 19.68          & 82.49             \\
    w/o Recon.                          & 47.57          & 2.48          & 20.49          & 83.56             \\
    w/o CMAD                             & 53.39          & 3.12          & 19.55          & 81.94             \\
    w/o ISD                              & 52.84          & 3.06          & 19.67          & 82.23             \\
    w/o WPA                              & 52.33          & 3.10          & 19.63          & 82.58             \\\bottomrule
    \end{tabular}
    \caption{Ablation study. 
    `G-Adv.', `L-Adv.' and `Adv.' denote global, local and both adversarial losses, respectively.
    `Recon.', `CMAD', `ISD' and `WPA' indicate reconstruction loss, cross-modal alignment distillation, in-sample distillation objectives and word patch alignment, respectively.}
    \label{tab:abl}
\end{table}
We conduct an ablation study to investigate the effectiveness of each component of our approach and the results are reported in Table \ref{tab:abl}. We first investigate the impact of the adversarial learning by removing global adversarial loss, local adversarial loss and both adversarial losses and find that the performance drops. The reason is that adversarial objectives focus on high-level features, which can effectively fill the missing region on complicated structures. Next, we test our method without reconstruction loss, which also decreases scores on all metrics. It demonstrates that image reconstruction plays an essential role in model training. Finally, we compare our method with the baseline without cross-modal alignment distillation and in-sample distillation, and the performance drops significantly. It indicates that the CMAD and ISD objectives can enhance the cross-modal alignment for text-guided image inpainting.

\subsection{Analysis}
\subsubsection{Impact of Cross-Modal Alignment}
\begin{figure}[t]
    \centering
    \begin{center}
    \includegraphics[width=0.9\linewidth]{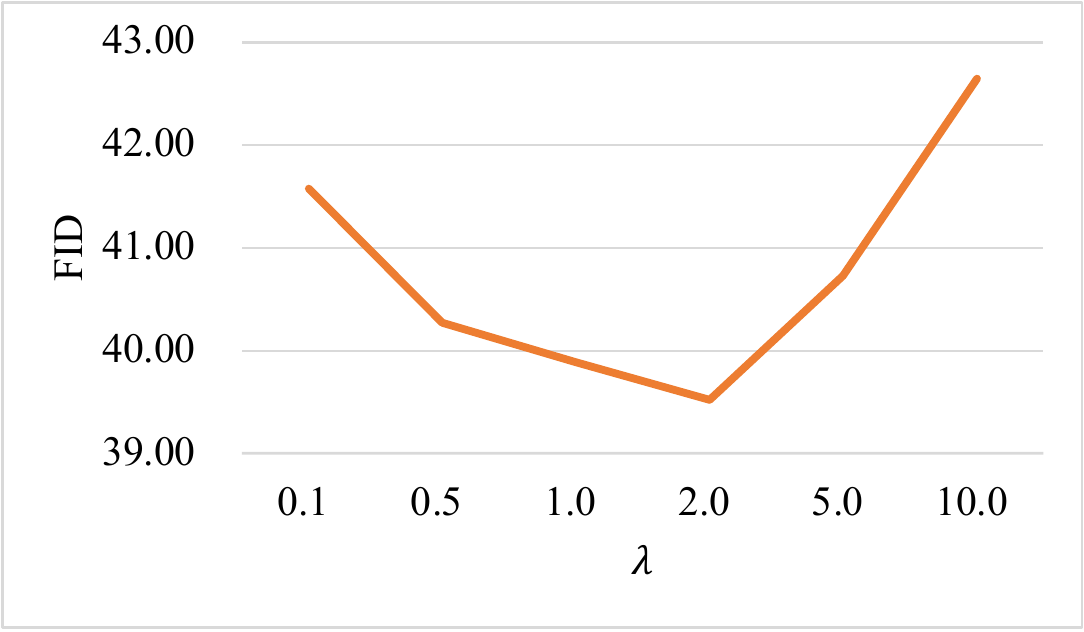}
    \end{center}
    \caption{Performance of our model with different trade-off parameters $\lambda$.}
    \label{fig:tradeoff}
\end{figure}
\begin{figure}[t]
    \centering
    \begin{center}
    \includegraphics[width=\linewidth]{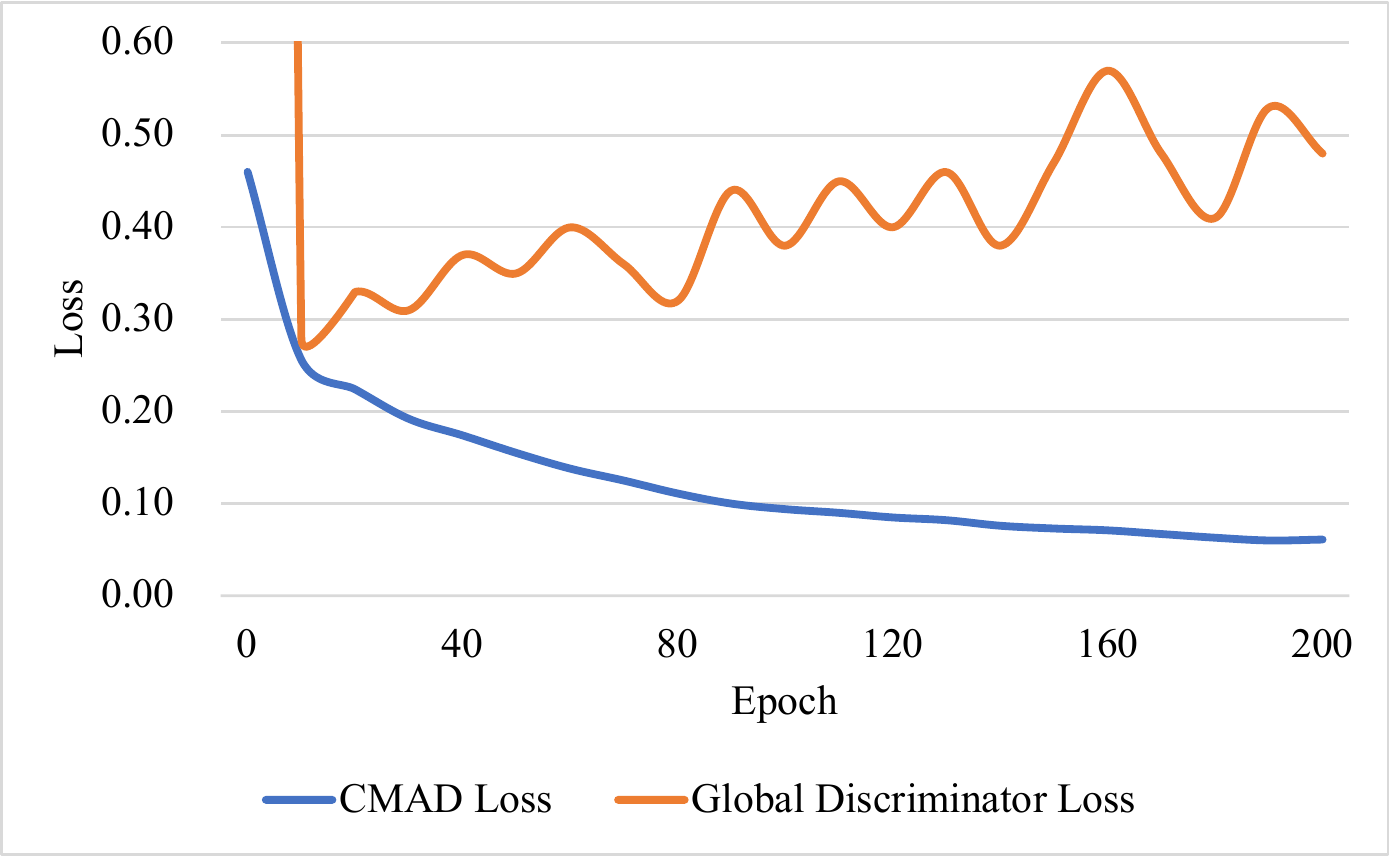}
    \end{center}
    \caption{The adversarial loss and cross-modal alignment distillation loss on the training set w.r.t different epochs in the training phase.}
    \label{fig:loss}
\end{figure}
To assess the impact of cross-modal alignment for our method, we set different trade-off parameters $\lambda$. Specifically, during model training, we set a different $\lambda$ for the loss function in Eq.\ref{equ:loss}. The performance of our method in different $\lambda$ is reported in Figure~\ref{fig:tradeoff}. It is observed that with the $\lambda$ increased, FID first drops and then rises, indicating that our method's performance first rises and then drops. The results demonstrate that the cross-modal alignment objective is crucial for our method, but overemphasizing the objective may hinder model learning.

\subsubsection{Deep Dive into Cross-Modal Alignment}
We take a deep dive into the impact of cross-modal alignment. In our method, adversarial learning involves a global discriminator to distinguish whether the generated image is original or generated, and the cross-modal alignment objective is to guide visual priors reconstruction. Their loss values, i.e., $\ell_{\mathrm{G-adv,D}}$ and $\ell_{\mathrm{CMAD}}$, are plotted in Figure~\ref{fig:loss}. We can see that the loss of cross-modal alignment objective quickly drops and then slowly drops, indicating that the model gets good performance on visual priors reconstruction. Meanwhile, we can observe that the adversarial loss of the discriminator quickly drops and then slowly goes up, demonstrating that we can see that the classification loss of the discriminator gets good performance and then is fooled later by the image generated from the image generator. Therefore, it shows that the cross-modal alignment objective supports the adversarial learning very well.

\subsubsection{User Study}
\begin{table}[t]\small
    \centering
    \begin{tabular}{lcc}\toprule
    \textbf{Method} & \multicolumn{1}{l}{\textbf{Naturalness}} & \multicolumn{1}{l}{\textbf{Semantic Consistency}} \\\midrule
    CMA             & 1.44                                     & 1.46                                              \\
    TDANet          & 2.39                                     & 2.52                                              \\
    ALMR            & 2.77                                     & 2.64                                              \\
    CSA             & 3.40                                     & 3.38                                              \\\bottomrule
    \end{tabular}
    \caption{Numerical ranking score of user study. The lower the score, the better the performance.}
    \label{tab:user}
\end{table}
\begin{figure}[t]
    \centering
    \begin{center}
    \includegraphics[width=\linewidth]{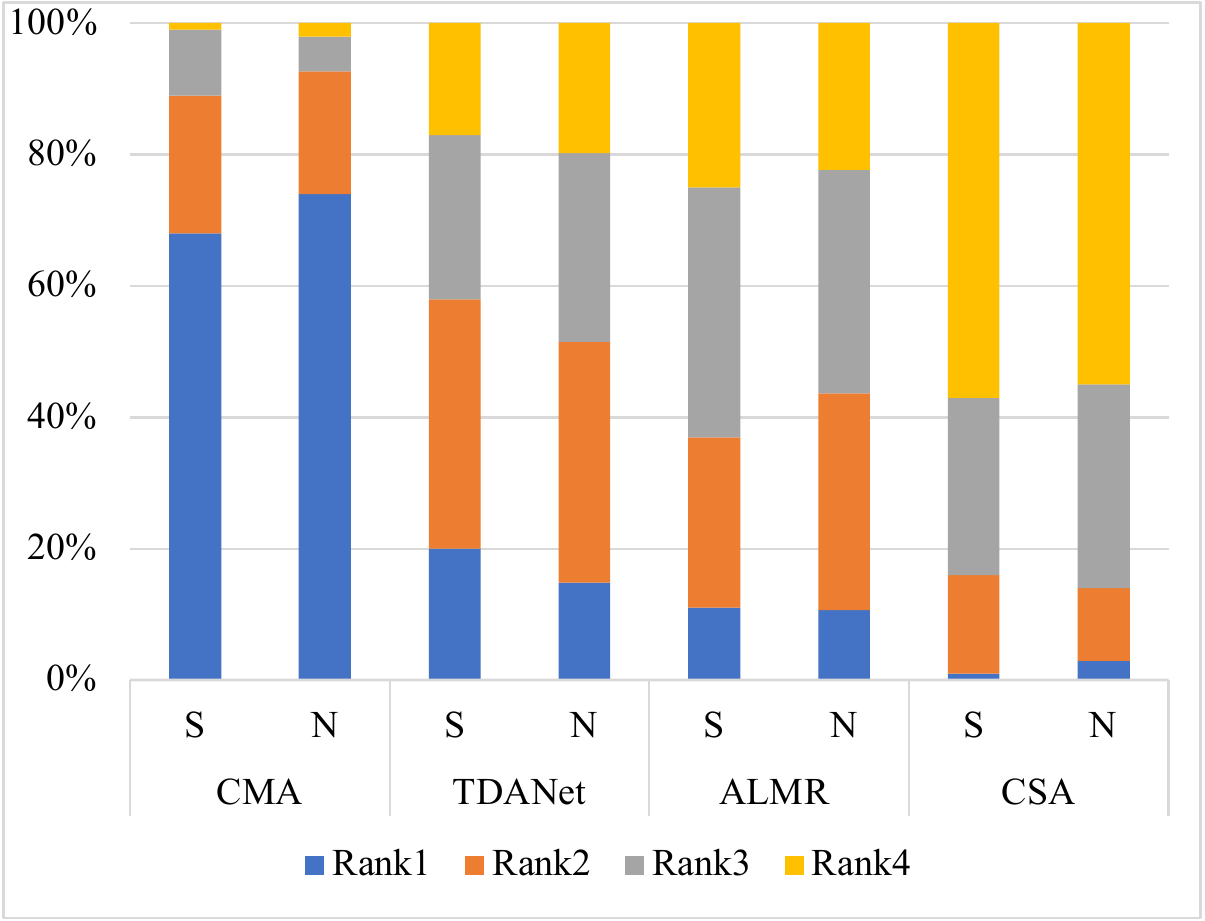}
    \end{center}
    \caption{Ranking score distribution of the user study. ``S'' means semantic consistency score, ``N'' means naturalness.}
    \label{fig:user}
\end{figure}
\begin{figure*}[t]
    \centering
    \begin{center}
    \includegraphics[width=\linewidth]{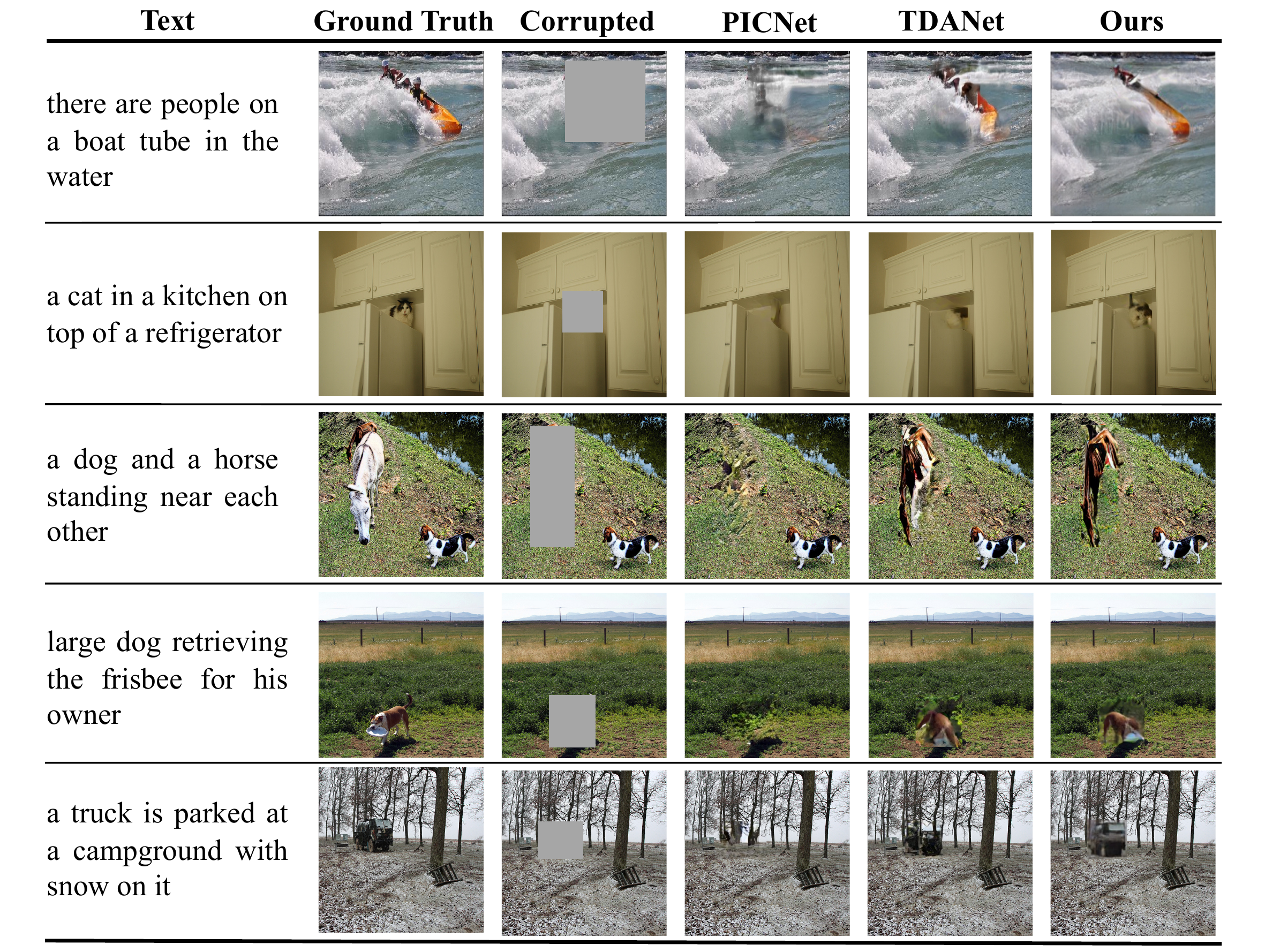}
    \end{center}
    \caption{Qualitative results randomly sampled from MSCOCO test set.}
    \label{fig:case}
\end{figure*}
Following \citet{Zhang20Text1}, we take a user study to quantify the qualitative comparison from the human perspective. We randomly collected 100 images in center masks from the MSCOCO test dataset. Each sample includes four generated images from CMA, PICNet, ALMR and CSA, respectively. These four images are randomly shuffled for five volunteers who rank images according to naturalness and the semantic consistency with the text description. Next, we computed the average ranking score as shown in Table~\ref{tab:user} and show their distribution in Figure~\ref{fig:user}. From the results, we can observe that our model outperforms other competitors in naturalness and semantic consistency. It demonstrates the effectiveness of our method from a qualitative perspective.

\subsection{Qualitative Results}
As demonstrated in Figure~\ref{fig:case}, we give qualitative comparison examples of our method and other methods. We can find that our model can generate more semantically plausible objects in the missing region. Firstly, comparing PICNet with TDANet and our method, extracting the semantic information from the text can improve the repair effect of the model in the missing area of the image. Secondly, we can observe that TDANet can generate preliminary results, such as the outlines and part content of "cat", "truck" and "people" in examples, but many details of the object are very unnatural. In contrast, our method can further generate the details of the object based on generating the outline and content of the object. It demonstrates that cross-modal alignment can effectively supplement missing object information.

\section{Related Work}
\subsection{Image Inpainting}
Image inpainting \cite{Marcelo00Image} aims to restore a damaged image, whose categories of approaches mainly are patch-based and deep learning-based. The patch-based methods \cite{barnes2009patchmatch,huang2014image} fill the holes through searching and pasting patches based on image known regions. \citet{huang2014image} propose a method using the mid-level structural cues to automatically guide the image inpainting. However, these methods are not effective to fill in the missing region on complicated structures, due to the focus on low-level features. To address the limitation of existing patch-based methods, there has been growing interest in deep learning-based methods \cite{pathak2016context,iizuka2017globally,ren2019structureflow}. The Context Encoder (CE) is proposed by \cite{pathak2016context}, which uses the encoder-decoder architecture and the Generative Adversarial Networks (GAN) \cite{goodfellow2014generative} to learn image features. Although the CE improves the inpainting by the image features learning, it is not effective to tackle the visual artifacts and exhibits blurriness in the image recovered regions. For solving the aforementioned problems, \citet{iizuka2017globally} introduce the local and global discriminator for the image inpainting of arbitrary missing regions, which improves the local and global consistency of generated image. The StructureFlow \cite{ren2019structureflow} consisted of a structure reconstructor and a texture generator, which can focus on recovering global structures and synthesizing high-frequency details. Although the existing image inpainting methods can fill in the holes in the image, generating specific content in the missing region remains challenging, without any known information.

\subsection{Text-Guided Image Inpainting}
To address this problem, many text-guided image inpainting works are proposed \cite{Zhang20Text,Lin20MMFL}. In these works, the specific content in the missing area can be restored based on the given descriptive text. Existing text-guided image inpainting methods include two processes: semantic information extraction and multimodal fusion. The semantic information extraction aims to obtain semantic information which does not match the image, such as the dual multimodal attention mechanism \cite{Zhang20Text1}. However, these methods are hard to work well in the image set with a variety of different objects. The reason is that these methods lack fine-grained alignment knowledge of texts and images to guide the fusion of cross-modal information in multimodal interactions. Besides, their fusion modules lack powerful cross-modal reasoning capabilities.

\subsection{Vision-Language Pre-training (VLP)}
Motivated by the success of the language/vision pre-trained model \cite{Devlin19BERT,Dosovitskiy21Image,Zhou22ClarET}, there is a surging interest in developing a pre-trained model for multiple modalities (e.g., vision and language) \cite{Chen20UNITER,Radford21Learning,Kim21ViLT,Zhou22Sketch}. For example, a pioneering work CLIP \cite{Radford21Learning} employs contrastive learning to predict whether matching between image and text and shows its powerful capability in many downstream tasks. UNITER \cite{Chen20UNITER} and UNIMO \cite{Chen20UNITER} employ an object detector (e.g., Faster R-CNN \cite{Ren17Faster}) to capture vision features, and a multi-layer transformer \cite{vaswani2017attention} is used to joint learn vision features and text features. \citet{Kim21ViLT} discuss different taxonomy of vision-and-language models and propose ViLT, a pre-trained model more focused on modeling modality interactions. In addition, ViLT totally discards convolutional visual features and adopts vision transformers. 

\section{Conclusion}
In this work, we explore a novel CMA model for text-guided image inpainting. In the CMA model, we integrate a vision encoder and a text encoder into a vision-and-language encoder, which is different from previous works. The vision-and-language encoder allocates more computation on modeling modality interactions instead of visual encoding. In addition, we introduce two objectives to improve cross-modal alignment, dubbed cross-modal alignment and in-sample distillation. The cross-modal alignment objective guides the model to fuse cross-modal features. Experimental results demonstrate that the proposed model delivers new state-of-the-art performance, followed by further analyses to provide comprehensive insights.

\section*{Limitations}
Compared with previous text-guided image inpainting methods \cite{Zhang20Text,Lin20MMFL,Wu21Adversarial}, our method performs well for natural image datasets with a wide variety of objects. However, the recovery of our method for completely missing objects in images is not perfect. The reason is that the size of our model limits its capabilities. The large image generation models (e.g., DALL$\cdot$E \cite{Ramesh21Zero}) show their powerful capability to generate a plausible image. Due to the limitation of computational resources, we could not train a large model of similar size to DALL$\cdot$E, which hinders verifying the effectiveness of our method on a large model.

\bibliography{ref}

\begin{thebibliography}{39}
\expandafter\ifx\csname natexlab\endcsname\relax\def\natexlab#1{#1}\fi

\bibitem[{Arjovsky et~al.(2017)Arjovsky, Chintala, and
  Bottou}]{arjovsky2017wasserstein}
Martin Arjovsky, Soumith Chintala, and L{\'e}on Bottou. 2017.
\newblock Wasserstein generative adversarial networks.
\newblock In \emph{International conference on machine learning}, pages
  214--223.

\bibitem[{Barnes et~al.(2009)Barnes, Shechtman, Finkelstein, and
  Goldman}]{barnes2009patchmatch}
Connelly Barnes, Eli Shechtman, Adam Finkelstein, and Dan~B Goldman. 2009.
\newblock Patchmatch: A randomized correspondence algorithm for structural
  image editing.
\newblock In \emph{ACM Transactions on Graphics (ToG)}, volume~28, page~24.
  ACM.

\bibitem[{Bertalm{\'{\i}}o et~al.(2000)Bertalm{\'{\i}}o, Sapiro, Caselles, and
  Ballester}]{Marcelo00Image}
Marcelo Bertalm{\'{\i}}o, Guillermo Sapiro, Vicent Caselles, and Coloma
  Ballester. 2000.
\newblock \href {https://doi.org/10.1145/344779.344972} {Image inpainting}.
\newblock In \emph{Proceedings of the 27th Annual Conference on Computer
  Graphics and Interactive Techniques, {SIGGRAPH} 2000, New Orleans, LA, USA,
  July 23-28, 2000}, pages 417--424. {ACM}.

\bibitem[{Chen et~al.(2020)Chen, Li, Yu, Kholy, Ahmed, Gan, Cheng, and
  Liu}]{Chen20UNITER}
Yen{-}Chun Chen, Linjie Li, Licheng Yu, Ahmed~El Kholy, Faisal Ahmed, Zhe Gan,
  Yu~Cheng, and Jingjing Liu. 2020.
\newblock \href {https://doi.org/10.1007/978-3-030-58577-8\_7} {{UNITER:}
  universal image-text representation learning}.
\newblock In \emph{Computer Vision - {ECCV} 2020 - 16th European Conference,
  Glasgow, UK, August 23-28, 2020, Proceedings, Part {XXX}}, volume 12375 of
  \emph{Lecture Notes in Computer Science}, pages 104--120. Springer.

\bibitem[{Devlin et~al.(2019)Devlin, Chang, Lee, and Toutanova}]{Devlin19BERT}
Jacob Devlin, Ming{-}Wei Chang, Kenton Lee, and Kristina Toutanova. 2019.
\newblock \href {https://doi.org/10.18653/v1/n19-1423} {{BERT:} pre-training of
  deep bidirectional transformers for language understanding}.
\newblock In \emph{Proceedings of the 2019 Conference of the North American
  Chapter of the Association for Computational Linguistics: Human Language
  Technologies, {NAACL-HLT} 2019, Minneapolis, MN, USA, June 2-7, 2019, Volume
  1 (Long and Short Papers)}, pages 4171--4186. Association for Computational
  Linguistics.

\bibitem[{Dosovitskiy et~al.(2021)Dosovitskiy, Beyer, Kolesnikov, Weissenborn,
  Zhai, Unterthiner, Dehghani, Minderer, Heigold, Gelly, Uszkoreit, and
  Houlsby}]{Dosovitskiy21Image}
Alexey Dosovitskiy, Lucas Beyer, Alexander Kolesnikov, Dirk Weissenborn,
  Xiaohua Zhai, Thomas Unterthiner, Mostafa Dehghani, Matthias Minderer, Georg
  Heigold, Sylvain Gelly, Jakob Uszkoreit, and Neil Houlsby. 2021.
\newblock \href {https://openreview.net/forum?id=YicbFdNTTy} {An image is worth
  16x16 words: Transformers for image recognition at scale}.
\newblock In \emph{9th International Conference on Learning Representations,
  {ICLR} 2021, Virtual Event, Austria, May 3-7, 2021}. OpenReview.net.

\bibitem[{Fardo et~al.(2016)Fardo, Conforto, de~Oliveira, and
  Rodrigues}]{Fardo16PSNR}
Fernando~A. Fardo, Victor~H. Conforto, Francisco~C. de~Oliveira, and Paulo~S.
  Rodrigues. 2016.
\newblock \href {http://arxiv.org/abs/1605.07116} {A formal evaluation of
  {PSNR} as quality measurement parameter for image segmentation algorithms}.
\newblock \emph{CoRR}, abs/1605.07116.

\bibitem[{Goodfellow et~al.(2014)Goodfellow, Pouget{-}Abadie, Mirza, Xu,
  Warde{-}Farley, Ozair, Courville, and Bengio}]{goodfellow2014generative}
Ian~J. Goodfellow, Jean Pouget{-}Abadie, Mehdi Mirza, Bing Xu, David
  Warde{-}Farley, Sherjil Ozair, Aaron~C. Courville, and Yoshua Bengio. 2014.
\newblock Generative adversarial nets.
\newblock In \emph{Advances in Neural Information Processing Systems 27: Annual
  Conference on Neural Information Processing Systems 2014, December 8-13 2014,
  Montreal, Quebec, Canada}, pages 2672--2680.

\bibitem[{Guo et~al.(2021)Guo, Yang, and Huang}]{Guo21Image}
Xiefan Guo, Hongyu Yang, and Di~Huang. 2021.
\newblock \href {https://doi.org/10.1109/ICCV48922.2021.01387} {Image
  inpainting via conditional texture and structure dual generation}.
\newblock In \emph{2021 {IEEE/CVF} International Conference on Computer Vision,
  {ICCV} 2021, Montreal, QC, Canada, October 10-17, 2021}, pages 14114--14123.
  {IEEE}.

\bibitem[{He et~al.(2016)He, Zhang, Ren, and Sun}]{He16Deep}
Kaiming He, Xiangyu Zhang, Shaoqing Ren, and Jian Sun. 2016.
\newblock \href {https://doi.org/10.1109/CVPR.2016.90} {Deep residual learning
  for image recognition}.
\newblock In \emph{2016 {IEEE} Conference on Computer Vision and Pattern
  Recognition, {CVPR} 2016, Las Vegas, NV, USA, June 27-30, 2016}, pages
  770--778. {IEEE} Computer Society.

\bibitem[{Heusel et~al.(2017)Heusel, Ramsauer, Unterthiner, Nessler, and
  Hochreiter}]{Heusel17GANs}
Martin Heusel, Hubert Ramsauer, Thomas Unterthiner, Bernhard Nessler, and Sepp
  Hochreiter. 2017.
\newblock \href
  {https://proceedings.neurips.cc/paper/2017/hash/8a1d694707eb0fefe65871369074926d-Abstract.html}
  {Gans trained by a two time-scale update rule converge to a local nash
  equilibrium}.
\newblock In \emph{Advances in Neural Information Processing Systems 30: Annual
  Conference on Neural Information Processing Systems 2017, December 4-9, 2017,
  Long Beach, CA, {USA}}, pages 6626--6637.

\bibitem[{Huang et~al.(2014)Huang, Kang, Ahuja, and Kopf}]{huang2014image}
Jia-Bin Huang, Sing~Bing Kang, Narendra Ahuja, and Johannes Kopf. 2014.
\newblock Image completion using planar structure guidance.
\newblock \emph{ACM Transactions on graphics (TOG)}, 33(4):129.

\bibitem[{Iizuka et~al.(2017)Iizuka, Simo-Serra, and
  Ishikawa}]{iizuka2017globally}
Satoshi Iizuka, Edgar Simo-Serra, and Hiroshi Ishikawa. 2017.
\newblock Globally and locally consistent image completion.
\newblock \emph{ACM Transactions on Graphics (ToG)}, 36(4):107.

\bibitem[{Kim et~al.(2021)Kim, Son, and Kim}]{Kim21ViLT}
Wonjae Kim, Bokyung Son, and Ildoo Kim. 2021.
\newblock \href {http://proceedings.mlr.press/v139/kim21k.html} {Vilt:
  Vision-and-language transformer without convolution or region supervision}.
\newblock In \emph{Proceedings of the 38th International Conference on Machine
  Learning, {ICML} 2021, 18-24 July 2021, Virtual Event}, volume 139 of
  \emph{Proceedings of Machine Learning Research}, pages 5583--5594. {PMLR}.

\bibitem[{Kingma and Ba(2015)}]{kingma2014adam}
Diederik~P. Kingma and Jimmy Ba. 2015.
\newblock Adam: {A} method for stochastic optimization.
\newblock In \emph{3rd International Conference on Learning Representations,
  {ICLR} 2015, San Diego, CA, USA, May 7-9, 2015, Conference Track
  Proceedings}.

\bibitem[{Kirillov et~al.(2020)Kirillov, Wu, He, and
  Girshick}]{Kirillov20PointRend}
Alexander Kirillov, Yuxin Wu, Kaiming He, and Ross~B. Girshick. 2020.
\newblock \href {https://doi.org/10.1109/CVPR42600.2020.00982} {Pointrend:
  Image segmentation as rendering}.
\newblock In \emph{2020 {IEEE/CVF} Conference on Computer Vision and Pattern
  Recognition, {CVPR} 2020, Seattle, WA, USA, June 13-19, 2020}, pages
  9796--9805. Computer Vision Foundation / {IEEE}.

\bibitem[{Lin et~al.(2020)Lin, Yan, Li, and Tan}]{Lin20MMFL}
Qing Lin, Bo~Yan, Jichun Li, and Weimin Tan. 2020.
\newblock \href {https://doi.org/10.1145/3394171.3413982} {{MMFL:} multimodal
  fusion learning for text-guided image inpainting}.
\newblock In \emph{{MM} '20: The 28th {ACM} International Conference on
  Multimedia, Virtual Event / Seattle, WA, USA, October 12-16, 2020}, pages
  1094--1102. {ACM}.

\bibitem[{Lin et~al.(2014)Lin, Maire, Belongie, Hays, Perona, Ramanan,
  Doll{\'{a}}r, and Zitnick}]{lin2014microsoft}
Tsung{-}Yi Lin, Michael Maire, Serge~J. Belongie, James Hays, Pietro Perona,
  Deva Ramanan, Piotr Doll{\'{a}}r, and C.~Lawrence Zitnick. 2014.
\newblock Microsoft {COCO:} common objects in context.
\newblock In \emph{Computer Vision - {ECCV} 2014 - 13th European Conference,
  Zurich, Switzerland, September 6-12, 2014, Proceedings, Part {V}}, volume
  8693 of \emph{Lecture Notes in Computer Science}, pages 740--755. Springer.

\bibitem[{Liu et~al.(2019)Liu, Jiang, Xiao, and Yang}]{Liu19Coherent}
Hongyu Liu, Bin Jiang, Yi~Xiao, and Chao Yang. 2019.
\newblock \href {https://doi.org/10.1109/ICCV.2019.00427} {Coherent semantic
  attention for image inpainting}.
\newblock In \emph{2019 {IEEE/CVF} International Conference on Computer Vision,
  {ICCV} 2019, Seoul, Korea (South), October 27 - November 2, 2019}, pages
  4169--4178. {IEEE}.

\bibitem[{Pathak et~al.(2016)Pathak, Kr{\"{a}}henb{\"{u}}hl, Donahue, Darrell,
  and Efros}]{pathak2016context}
Deepak Pathak, Philipp Kr{\"{a}}henb{\"{u}}hl, Jeff Donahue, Trevor Darrell,
  and Alexei~A. Efros. 2016.
\newblock Context encoders: Feature learning by inpainting.
\newblock In \emph{2016 {IEEE} Conference on Computer Vision and Pattern
  Recognition, {CVPR} 2016, Las Vegas, NV, USA, June 27-30, 2016}, pages
  2536--2544. {IEEE} Computer Society.

\bibitem[{Plummer et~al.(2017)Plummer, Wang, Cervantes, Caicedo, Hockenmaier,
  and Lazebnik}]{Plummer17Flickr30k}
Bryan~A. Plummer, Liwei Wang, Chris~M. Cervantes, Juan~C. Caicedo, Julia
  Hockenmaier, and Svetlana Lazebnik. 2017.
\newblock \href {https://doi.org/10.1007/s11263-016-0965-7} {Flickr30k
  entities: Collecting region-to-phrase correspondences for richer
  image-to-sentence models}.
\newblock \emph{Int. J. Comput. Vis.}, 123(1):74--93.

\bibitem[{Radford et~al.(2021)Radford, Kim, Hallacy, Ramesh, Goh, Agarwal,
  Sastry, Askell, Mishkin, Clark, Krueger, and Sutskever}]{Radford21Learning}
Alec Radford, Jong~Wook Kim, Chris Hallacy, Aditya Ramesh, Gabriel Goh,
  Sandhini Agarwal, Girish Sastry, Amanda Askell, Pamela Mishkin, Jack Clark,
  Gretchen Krueger, and Ilya Sutskever. 2021.
\newblock \href {http://proceedings.mlr.press/v139/radford21a.html} {Learning
  transferable visual models from natural language supervision}.
\newblock In \emph{Proceedings of the 38th International Conference on Machine
  Learning, {ICML} 2021, 18-24 July 2021, Virtual Event}, volume 139 of
  \emph{Proceedings of Machine Learning Research}, pages 8748--8763. {PMLR}.

\bibitem[{Ramesh et~al.(2021)Ramesh, Pavlov, Goh, Gray, Voss, Radford, Chen,
  and Sutskever}]{Ramesh21Zero}
Aditya Ramesh, Mikhail Pavlov, Gabriel Goh, Scott Gray, Chelsea Voss, Alec
  Radford, Mark Chen, and Ilya Sutskever. 2021.
\newblock \href {http://proceedings.mlr.press/v139/ramesh21a.html} {Zero-shot
  text-to-image generation}.
\newblock In \emph{Proceedings of the 38th International Conference on Machine
  Learning, {ICML} 2021, 18-24 July 2021, Virtual Event}, volume 139 of
  \emph{Proceedings of Machine Learning Research}, pages 8821--8831. {PMLR}.

\bibitem[{Ren et~al.(2017)Ren, He, Girshick, and Sun}]{Ren17Faster}
Shaoqing Ren, Kaiming He, Ross~B. Girshick, and Jian Sun. 2017.
\newblock \href {https://doi.org/10.1109/TPAMI.2016.2577031} {Faster {R-CNN:}
  towards real-time object detection with region proposal networks}.
\newblock \emph{{IEEE} Trans. Pattern Anal. Mach. Intell.}, 39(6):1137--1149.

\bibitem[{Ren et~al.(2019)Ren, Yu, Zhang, Li, Liu, and
  Li}]{ren2019structureflow}
Yurui Ren, Xiaoming Yu, Ruonan Zhang, Thomas~H. Li, Shan Liu, and Ge~Li. 2019.
\newblock Structureflow: Image inpainting via structure-aware appearance flow.
\newblock In \emph{2019 {IEEE/CVF} International Conference on Computer Vision,
  {ICCV} 2019, Seoul, Korea (South), October 27 - November 2, 2019}, pages
  181--190. {IEEE}.

\bibitem[{Rudin et~al.(1992)Rudin, Osher, and Fatemi}]{rudin1992nonlinear}
Leonid~I Rudin, Stanley Osher, and Emad Fatemi. 1992.
\newblock Nonlinear total variation based noise removal algorithms.
\newblock \emph{Physica D: nonlinear phenomena}, 60(1-4):259--268.

\bibitem[{Vaswani et~al.(2017)Vaswani, Shazeer, Parmar, Uszkoreit, Jones,
  Gomez, Kaiser, and Polosukhin}]{vaswani2017attention}
Ashish Vaswani, Noam Shazeer, Niki Parmar, Jakob Uszkoreit, Llion Jones,
  Aidan~N. Gomez, Lukasz Kaiser, and Illia Polosukhin. 2017.
\newblock Attention is all you need.
\newblock In \emph{Advances in Neural Information Processing Systems 30: Annual
  Conference on Neural Information Processing Systems 2017, 4-9 December 2017,
  Long Beach, CA, {USA}}, pages 5998--6008.

\bibitem[{Wah et~al.(2011)Wah, Branson, Welinder, Perona, and
  Belongie}]{wah2011caltech}
Catherine Wah, Steve Branson, Peter Welinder, Pietro Perona, and Serge
  Belongie. 2011.
\newblock The caltech-ucsd birds-200-2011 dataset.

\bibitem[{Wan et~al.(2021)Wan, Zhang, Chen, and Liao}]{Wan21High}
Ziyu Wan, Jingbo Zhang, Dongdong Chen, and Jing Liao. 2021.
\newblock \href {https://doi.org/10.1109/ICCV48922.2021.00465} {High-fidelity
  pluralistic image completion with transformers}.
\newblock In \emph{2021 {IEEE/CVF} International Conference on Computer Vision,
  {ICCV} 2021, Montreal, QC, Canada, October 10-17, 2021}, pages 4672--4681.
  {IEEE}.

\bibitem[{Wang et~al.(2004)Wang, Bovik, Sheikh, and Simoncelli}]{Wang04Image}
Zhou Wang, Alan~C. Bovik, Hamid~R. Sheikh, and Eero~P. Simoncelli. 2004.
\newblock \href {https://doi.org/10.1109/TIP.2003.819861} {Image quality
  assessment: from error visibility to structural similarity}.
\newblock \emph{{IEEE} Trans. Image Process.}, 13(4):600--612.

\bibitem[{Wang et~al.(2021)Wang, Yu, Yu, Dai, Tsvetkov, and Cao}]{Wang21SimVLM}
Zirui Wang, Jiahui Yu, Adams~Wei Yu, Zihang Dai, Yulia Tsvetkov, and Yuan Cao.
  2021.
\newblock \href {http://arxiv.org/abs/2108.10904} {Simvlm: Simple visual
  language model pretraining with weak supervision}.
\newblock \emph{CoRR}, abs/2108.10904.

\bibitem[{Wu et~al.(2021)Wu, Xie, Zeng, Yang, Yu, Li, and
  Liu}]{Wu21Adversarial}
Xingcai Wu, Yucheng Xie, Jiaqi Zeng, Zhenguo Yang, Yi~Yu, Qing Li, and Wenyin
  Liu. 2021.
\newblock \href {https://doi.org/10.1145/3474085.3475506} {Adversarial learning
  with mask reconstruction for text-guided image inpainting}.
\newblock In \emph{{MM} '21: {ACM} Multimedia Conference, Virtual Event, China,
  October 20 - 24, 2021}, pages 3464--3472. {ACM}.

\bibitem[{Zhang et~al.(2020{\natexlab{a}})Zhang, Chen, Hu, and
  Jiang}]{Zhang20Text1}
Lisai Zhang, Qingcai Chen, Baotian Hu, and Shuoran Jiang. 2020{\natexlab{a}}.
\newblock \href {https://doi.org/10.1145/3394171.3414017} {Text-guided neural
  image inpainting}.
\newblock In \emph{{MM} '20: The 28th {ACM} International Conference on
  Multimedia, Virtual Event / Seattle, WA, USA, October 12-16, 2020}, pages
  1302--1310. {ACM}.

\bibitem[{Zhang et~al.(2020{\natexlab{b}})Zhang, Zhao, Zhang, Huai, and
  Yuan}]{Zhang20Text}
Zijian Zhang, Zhou Zhao, Zhu Zhang, Baoxing Huai, and Jing Yuan.
  2020{\natexlab{b}}.
\newblock \href {https://doi.org/10.1145/3394171.3413939} {Text-guided image
  inpainting}.
\newblock In \emph{{MM} '20: The 28th {ACM} International Conference on
  Multimedia, Virtual Event / Seattle, WA, USA, October 12-16, 2020}, pages
  4079--4087. {ACM}.

\bibitem[{Zheng et~al.(2019)Zheng, Cham, and Cai}]{Zheng19Pluralistic}
Chuanxia Zheng, Tat{-}Jen Cham, and Jianfei Cai. 2019.
\newblock \href {https://doi.org/10.1109/CVPR.2019.00153} {Pluralistic image
  completion}.
\newblock In \emph{{IEEE} Conference on Computer Vision and Pattern
  Recognition, {CVPR} 2019, Long Beach, CA, USA, June 16-20, 2019}, pages
  1438--1447. Computer Vision Foundation / {IEEE}.

\bibitem[{Zhou(2022)}]{Zhou22Sketch}
Yucheng Zhou. 2022.
\newblock \href {https://doi.org/10.1109/ICASSP43922.2022.9746558} {Sketch
  storytelling}.
\newblock In \emph{{IEEE} International Conference on Acoustics, Speech and
  Signal Processing, {ICASSP} 2022, Virtual and Singapore, 23-27 May 2022},
  pages 4748--4752. {IEEE}.

\bibitem[{Zhou et~al.(2022{\natexlab{a}})Zhou, Geng, Shen, Long, and
  Jiang}]{Zhou22EventBERT}
Yucheng Zhou, Xiubo Geng, Tao Shen, Guodong Long, and Daxin Jiang.
  2022{\natexlab{a}}.
\newblock \href {https://doi.org/10.1145/3485447.3511928} {Eventbert: {A}
  pre-trained model for event correlation reasoning}.
\newblock In \emph{{WWW} '22: The {ACM} Web Conference 2022, Virtual Event,
  Lyon, France, April 25 - 29, 2022}, pages 850--859. {ACM}.

\bibitem[{Zhou et~al.(2022{\natexlab{b}})Zhou, Shen, Geng, Long, and
  Jiang}]{Zhou22ClarET}
Yucheng Zhou, Tao Shen, Xiubo Geng, Guodong Long, and Daxin Jiang.
  2022{\natexlab{b}}.
\newblock \href {https://doi.org/10.18653/v1/2022.acl-long.183} {Claret:
  Pre-training a correlation-aware context-to-event transformer for
  event-centric generation and classification}.
\newblock In \emph{Proceedings of the 60th Annual Meeting of the Association
  for Computational Linguistics (Volume 1: Long Papers), {ACL} 2022, Dublin,
  Ireland, May 22-27, 2022}, pages 2559--2575. Association for Computational
  Linguistics.

\bibitem[{Zhu et~al.(2020)Zhu, Shen, Zhao, and Zhou}]{Zhu20In}
Jiapeng Zhu, Yujun Shen, Deli Zhao, and Bolei Zhou. 2020.
\newblock \href {https://doi.org/10.1007/978-3-030-58520-4\_35} {In-domain
  {GAN} inversion for real image editing}.
\newblock In \emph{Computer Vision - {ECCV} 2020 - 16th European Conference,
  Glasgow, UK, August 23-28, 2020, Proceedings, Part {XVII}}, volume 12362 of
  \emph{Lecture Notes in Computer Science}, pages 592--608. Springer.

\end{thebibliography}
\bibliographystyle{acl_natbib}

\appendix

\end{document}